\documentclass[10pt, a4paper]{article}

\usepackage{lrec-coling2024}

\usepackage{inconsolata}
\usepackage{amsmath}
\usepackage{amsfonts}
\usepackage{amssymb}
\usepackage{booktabs}
\usepackage{csquotes}
\usepackage{listings}
\usepackage{float}
\usepackage{multirow}
\usepackage{multicol}
\usepackage{verbatim}
\usepackage[super]{nth}
\usepackage[capitalize,nameinlink]{cleveref}
\usepackage{tikz} 
\usetikzlibrary{positioning,calc}
\usepackage{mathastext}
\usepackage{caption}
\usepackage{subcaption}

\definecolor{lavender}{RGB}{208,180,240}

\newcommand{\eg}{\textit{e.g.}}
\newcommand{\ie}{\textit{i.e.}}

\newcommand{\code}[1]{\texttt{#1}}
\newcommand{\nop}[1]{}

\title{Nostra Domina at EvaLatin 2024:\\Improving Latin Polarity Detection through Data Augmentation}

\name{Stephen Bothwell, Abigail Swenor, David Chiang} 

\address{University of Notre Dame \\
         Notre Dame, Indiana, USA \\
         \{sbothwel, aswenor, dchiang\}@nd.edu \\}

\abstract{
This paper describes submissions from the team Nostra Domina to the EvaLatin 2024 shared task of emotion polarity detection. Given the low-resource environment of Latin and the complexity of sentiment in rhetorical genres like poetry, we augmented the available data through automatic polarity annotation. We present two methods for doing so on the basis of the $k$-means algorithm, and we employ a variety of Latin large language models (LLMs) in a neural architecture to better capture the underlying contextual sentiment representations. Our best approach achieved the second highest macro-averaged Macro-F\textsubscript{1} score on the shared task's test set.
 \\ \newline  }

\begin{document}

\maketitleabstract

\section{Introduction}
\label{sec:introduction}

Emotion polarity detection is a variant on the common NLP task of sentiment analysis. Usual applications of this task tend to be on reviews---for example, about movies \citeplanguageresource{maasLearningWordVectors2011,socherRecursiveDeepModels2013} or products \citeplanguageresource{blitzerBiographiesBollywoodBoomboxes2007}---where providing an opinion is the author's goal. Few works have extended this task to less direct modalities of sentiment, like poetry, and even fewer to ancient languages, like Latin \citeplanguageresource{chenBuildingSentimentLexicons2014,marleySentimentsNetworksLiterary2018,sprugnoliOdiAmoCreating2020,sprugnoliSentimentLatinPoetry2023}. Thus, the EvaLatin 2024 evaluation campaign's take on this task \cite{sprugnoliOverviewEvaLatin2024} tackles both an uncommon genre and a low-resource environment.

Motivated by the lack of sentiment resources, this work presents two methods for the automatic annotation of data: \textit{polarity coordinate} clustering, a novel specialization on $k$-means clustering, and Gaussian clustering. Furthermore, our work examines a variety of different Latin LLMs in a straightforward neural architecture through a hyperparameter search to determine their efficacy on the emotion polarity detection task. To our knowledge, we are the first outside of the original authors to explicitly apply some of these language models for Latin.\footnote{We make our data and code available at: \url{https://github.com/Mythologos/EvaLatin2024-NostraDomina}.} 

After we introduce the small set of pre-existing data for this task, we describe our clustering-based annotation methods (\Cref{sec:automatic-annotation}) and their results (\Cref{sec:annotation-results}). Then, we describe our neural architecture (\Cref{sec:modeling}) and the procedure used for model training and selection (\Cref{sec:experimental-design}). Finally, we go over our results for this task and investigate why our models performed as they did, with one achieving the second best macro-averaged Macro-F\textsubscript{1} score on EvaLatin's test set (\Cref{sec:results}). 


\section{Data}
\label{sec:data}

\begin{figure}
    \small
    \centering
    \includegraphics[width=.8\linewidth]{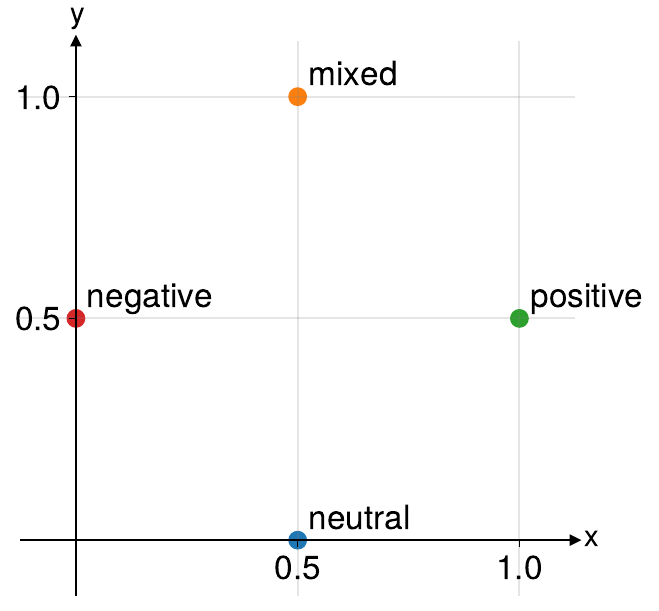}
    \vspace{-.25cm}
    \caption{The polarity coordinate plane. Points are all colored differently to represent their classes and are labeled accordingly. The $x$-axis and $y$-axis represent polarity and intensity, respectively.}
    \label{fig:polarity-coordinate-plane}
\end{figure}



Very little data exists for sentiment analysis in Latin. Until recently, only static representations of sentiment were available in sentiment lexica. To our knowledge, the first Latin sentiment lexicon was one automatically transferred to Latin based on English lexica and a large knowledge graph \citeplanguageresource{chenBuildingSentimentLexicons2014}. This was followed by two others. One was manually curated by a single author based on Stoic values in a study on Cicero \citeplanguageresource{marleySentimentsNetworksLiterary2018}. The other, called LatinAffectus, was created by multiple Latin experts and organized according to inter-annotator agreement \citeplanguageresource{sprugnoliOdiAmoCreating2020}; its most recent 
version, LatinAffectus-v4, was released for use in this shared task.

Following this, \citeauthor{sprugnoliSentimentLatinPoetry2023} released the first dataset for Latin sentiment analysis. This dataset, having the same classes as our shared task, covers a selection of Horace's \textit{Odes}---a staple of classical poetry. It contains 44 labeled sentences and has the class distribution given in \Cref{tab:dataset-distributions}. Although this dataset lays groundwork for future studies in Latin sentiment analysis, it is not large enough to train a traditional neural classifier. This is especially the case for a genre which indirectly conveys opinions: poetry frequently employs allusion (\eg, to contemporary circumstances) and rhetorical devices (\eg, metaphor, sarcasm) to make its points.

Given this lack of available training data, we investigate automatic annotation to approximate sentiment for Latin. Because of the variety of time periods, genres, and additional annotations covered by the Universal Dependency (UD) \citeplanguageresource{demarneffeUniversalDependencies2021} treebanks for Latin, we select each of the Perseus \citeplanguageresource{smithPerseusProjectDigital2000,bammanAncientGreekLatin2011}, PROIEL \citeplanguageresource{haugComputationalLinguisticIssues2009}, ITTB \citeplanguageresource{passarottiProjectIndexThomisticus2019}, LLCT \citeplanguageresource{cecchiniNewLatinTreebank2020}, and UDante \citeplanguageresource{cecchiniUDanteFirstSteps2020a} treebanks for this purpose. We also incorporate data from EvaLatin 2022 \citeplanguageresource{sprugnoliOverviewEvaLatin2022} and the Archimedes Latinus treebank \citeplanguageresource{fantoliLinguisticAnnotationNeoLatin2022}. 

\begin{table}[t]
    \small
    \centering
    \begin{tabular}{lrrrr}
        \toprule
        \multirow[t]{2}{3em}{\textbf{Dataset}} & \multicolumn{4}{c}{\textbf{Class}} \\
        & Positive & Negative & Neutral & Mixed \\
        \midrule
        \textit{Odes} & 20 & 12 & 3 & 9 \\
        \midrule[.1em]
        \midrule[.05em]
        PC & 10427 & 4114 & 57786 & 4178 \\
        Gaussian & 33473 & 14333 & 16861 & 11838 \\
        \midrule[.1em]
        \midrule[.05em]
        Horace & 20 & 55 & 8 & 15 \\
        Pontano & 48 & 18 & 10 & 22 \\
        Seneca & 7 & 81 & 2 & 13 \\
        \midrule[.01em]
        Total & 75 & 154 & 20 & 50 \\
        \bottomrule
    \end{tabular}
    \caption{Resource class distributions. The top, middle, and bottom sections (broken up by pairs of lines) concern pre-existing resources, new resources, and EvaLatin test subsets (or the total set), respectively. \enquote{PC} is Polarity Coordinate.}
    \label{tab:dataset-distributions}
\end{table}

\subsection{Automatic Annotation}
\label{sec:automatic-annotation}

In this section, we detail our data augmentation methods. Both methods relate to the $k$-means clustering algorithm, where central points---\textit{centroids}---are selected, and the distances between a data point and these centroids relate them in some way. 

\subsubsection{Polarity Coordinate (PC) Clustering}
\label{sec:polarity-coordinate-clustering}

The task of emotion polarity detection, for the available Latin sentiment data, categorizes each sentence into one of four classes: \textit{positive}, \textit{negative}, \textit{neutral}, and \textit{mixed} \citeplanguageresource{sprugnoliSentimentLatinPoetry2023}. This set of classes stems from the circumplex model of affect \cite{russellEvidenceThreefactorTheory1977,russellCircumplexModelAffect1980} in which emotions are plotted on a two-dimensional plane with the axes of pleasure-displeasure and arousal-sleep. Sentiment analysis works have often applied this theory with varying terminology \cite{tianPolarityIntensityTwo2018}. In our case, we use \textit{polarity} to refer to the \enquote{direction} of sentiment (\ie, pleasing or displeasing) and \textit{intensity} to refer to the \textit{magnitude} of the sentiment (\ie, aroused or inert). 

These definitions of polarity and intensity can be used to differentiate the four classes for our task. For a given sentence, if its polarity is definitively pleasing, then it is positive; if its polarity is definitively displeasing, then it is negative; if its polarity has both positive and negative elements and has high intensity, then it is mixed; and if it fits into none of these categories (\ie, there is no moderate intensity in either direction), then it is neutral. We employ this mapping to classify sentences via the $k$-means algorithm. To do so, we must determine the representation for our classes as centroids and our sentences as data points.

Following the idea of the circumplex model, we establish polarity and intensity on a coordinate plane. However, we map the space of these values between 0 and 1, meaning that the point (0.5, 0.5) represents a point of average polarity and intensity. This point is equidistant from each of the four designated class centroids, which we present in \Cref{fig:polarity-coordinate-plane}. Although the positive and negative classes have no innate relation to intensity, we assume that some intensity must exist for the polarity to be noticeable. Given these centroids, we then define a polarity coordinate $P$ for a sequence $\mathbf{x}$ as:
\begin{align}
    P_{\mathbf{x}} & = (\text{polarity}(\mathbf{x}), \text{intensity}(\mathbf{x})) \\
    \text{polarity}(\mathbf{x}) & = \left(\frac{1}{2|\mathbf{x}|} \sum_{i=1}^{|\mathbf{x}|} \text{score}(x_{i})\right) + \frac{1}{2} \\
    \text{intensity}(\mathbf{x}) & = \frac{1}{|\mathbf{x}|} \sum_{i=1}^{|\mathbf{x}|} |\text{score}(x_{i})|
\end{align}
and \code{score} outputs values between -1 and 1.

To classify sentences, we used  LatinAffectus-v4 as the crux of our scoring function. Each $x_{i} \in \mathbf{x}$ was searched in the lexicon. To search the lexicon, we used lemmata from the treebank sentences if they were available and the \code{LatinBackoffLemmatizer} from the Classical Language Toolkit (CLTK) as a backoff option \cite{johnsonClassicalLanguageToolkit2021}.\footnote{While not necessary for most treebank data, the Archimedes Latinus treebank \citeplanguageresource{fantoliLinguisticAnnotationNeoLatin2022} does not provide lemmata.} To prevent the impact of sentiment words from being diminished due to the fact that the majority of words were not found in the lexicon, we only used words in LatinAffectus-v4 to score each sentence. This meant that the \code{polarity} and \code{intensity} functions would receive a filtered $\mathbf{x'}$ rather than $\mathbf{x}$. Sentences with no lexical entries were deemed neutral.

Although this method was inspired by the task structure, we suspected that its outputs would be noisy, as it employed static sentiment representations. To account for the noise, we attempted to use the distances between a sentence and each centroid to our advantage. Suppose that we have a collection of distances $\mathbf{d}$. We normalized these distances $\mathbf{d}$; call this set of normalized distances $\mathbf{d}'$. Then, we calculated a value $\alpha$ for each sentence by subtracting $\min{(\mathbf{d}')}$ from $1$. This $\alpha$ serves as a confidence value for the given label. If the distance for a sentence and its label is low, then the sentence may be a stronger representative for that class and can aid more in the learning process.

With this in mind, we augmented the traditional cross-entropy loss function with a set of these $\alpha$ values, forming what we call the \textit{gold distance weighted cross-entropy} (GDW-CE) loss. Given predictions $\mathbf{Y}'$ and ground truth values $\mathbf{Y}$ (where $|\mathbf{Y}'| = |\mathbf{Y}| = N$), confidence values $\boldsymbol{\alpha}$, and the cross-entropy function $H$, the equation for this loss is:
\begin{equation}
    \textbf{GDW-CE}(\mathbf{Y'}, \mathbf{Y}, \boldsymbol{\alpha}) = \sum_{i=0}^{N} (\alpha_{i} * H(Y'_{i}, Y_{i}))
\end{equation}

\subsubsection{Gaussian Clustering}
\label{sec:gaussian-clustering}


Unlike $k$-means clustering, Gaussian clustering does not serve as an explicit classifier; instead, it outputs the probabilities for which a given data point is within each cluster. Naturally, however, we can take the cluster with the highest probability to be the label for any given data point. Once again, then, what remains is to establish how the class and sentence representations are derived.

To derive class representations, we trained a Gaussian Mixture Model (GMM) drawn from four distributions (\ie, classes) on the \textit{Odes} dataset \cite{sprugnoliSentimentLatinPoetry2023}. We fitted a GMM with the \code{scikit-learn} library \cite{virtanenSciPyFundamentalAlgorithms2020} via the expectation-maximization algorithm. To gather representations for each sentence, we computed sentence-level embeddings from the SPhilBERTa model \cite{riemenschneiderGraeciaCaptaFerum2023}, a pre-trained language model for English, Latin, and Ancient Greek based on the Sentence-BERT architecture \cite{reimersSentenceBERTSentenceEmbeddings2019}. We appended the polarity coordinate features described in \Cref{sec:polarity-coordinate-clustering} to these embeddings.

We performed a hyperparameter grid search to select the best GMM. Due to space considerations, we defer the details of this search to our repository. Because of the available data's size, trials were both trained and evaluated on the \textit{Odes} for their Macro-F\textsubscript{1} score; the best GMM scored $0.37$.


\subsubsection{Annotation Results}
\label{sec:annotation-results}

The outcomes of both annotation methods are provided in the middle of \Cref{tab:dataset-distributions}. The PC and Gaussian datasets have dissimilar distributions, preferring the neutral and positive classes, respectively. 

\section{Modeling}
\label{sec:modeling}

\begin{figure}[t]
    \small
    \centering 
    \begin{tikzpicture}[x=1cm,y=1cm]
        \begin{scope}[every path/.style={fill=lavender, very thick}, every text node part/.style={align=center}]
            \draw (-1,0.35) rectangle (1,1.10);
            \node at (0,.725) {\textsc{Embedding}};
            \node[anchor=west] at (1.5,.725) { \textbf{\{}\textsc{Latin BERT}, \textsc{LaBERTa}, \\ \textsc{PhilBERTa}, \textsc{mBERT}, \textsc{CANINE-C}, \\ \textsc{CANINE-S}, \textsc{SPhilBERTa}\textbf{\}} };
            \draw (-1,1.45) -- (-1.25,1.825) -- (-1,2.20) -- (1,2.20) -- (1.25,1.825) -- (1,1.45) -- cycle;
            \node at (0,1.825) {\textsc{Encoder}};
            \node[anchor=west] at (1.5,1.825) { \textbf{\{}\textsc{None}, \textsc{BiLSTM}, \textsc{Transformer}\textbf{\}} };
            \draw (-1,2.55) -- (-0.4,3.30) -- (0.4,3.30) -- (1,2.55) -- cycle;
            \node at (0,2.925) {\textsc{Linear}};
        \end{scope}
        \begin{scope}[->,>=latex,very thick]
            \draw (0,0) -- (0,0.35);
            \draw (0,1.10) -- (0,1.45);
            \draw (0,2.20) -- (0,2.55);
            \draw (0,3.30) -- (0,3.65);
        \end{scope}
    \end{tikzpicture}
    \vspace{-.5cm}
    \caption{Architectural options fixed across hyperparameter search trials. Shapes reflect the relative dimensionality of data throughout the network.}
    \label{fig:classifier-architecture}
\end{figure}
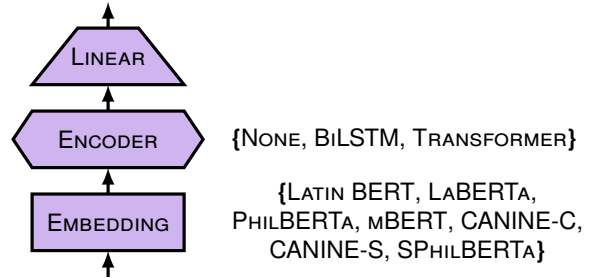

We apply a basic neural architecture for the emotion polarity detection task. As \Cref{fig:classifier-architecture} depicts, there are three main parts to this architecture: the embedding, encoder, and linear layers. 
For the embedding and encoder layers, we have alternatives for each which we examine in our experiments. 

For our embeddings, we use all known publicly-available encoder-based LMs containing Latin. 
Latin BERT \cite{bammanLatinBERTContextual2020}, LaBERTa and PhilBERTa \cite{riemenschneiderExploringLargeLanguage2023}, and SPhilBERTa \cite{riemenschneiderGraeciaCaptaFerum2023} are all either monolingual models (in the case of Latin BERT and LaBERTa) or classical trilingual models (PhilBERTa and SPhilBERTa). We also used the multilingual mBERT \cite{devlinBERTPretrainingDeep2019} and the character-based CANINE-C and CANINE-S \cite{clarkCaninePretrainingEfficient2022}, trained with character-based and subword-based losses, respectively. We froze embeddings during training to maintain their contextual representations.

For our encoders, we employ an identity transformation, a bidirectional LSTM (BiLSTM) \cite{gravesFramewisePhonemeClassification2005}, and a Transformer \cite{vaswaniAttentionAllYou2017}. For the BiLSTM, we concatenate the final hidden states for both directional LSTMs to provide the final state for classification. For the identity layer and the Transformer, we select the \code{[CLS]} token's representation.

\section{Experiments}
\label{sec:experiments}


\begin{table}[t]
    \small
    \centering
    \begin{tabular}{lrrr}
        \toprule
        \multirow[t]{2}{5em}{\textbf{Embedding}} & \multicolumn{3}{c}{\textbf{Encoder}} \\
        & Identity & LSTM & Transformer \\
        \midrule
        Latin BERT & 0.12$^{\dagger}$ & 0.21$^{*}$ & 0.12$^{\dagger}$ \\
        LaBERTa & 0.03$^{*}$ & 0.17$^{*}$ & 0.21$^{\dagger}$ \\
        PhilBERTa & 0.06$^{*}$ & 0.15$^{*}$ & 0.13$^{\dagger}$ \\
        mBERT & 0.07$^{\dagger}$ & 0.09$^{\dagger}$ & 0.08$^{\dagger}$ \\
        CANINE-C & 0.14$^{*}$ & 0.20$^{\dagger}$ & 0.03$^{*}$ \\
        CANINE-S & 0.08$^{\dagger}$ & 0.17$^{*}$ & 0.18$^{\dagger}$ \\
        SPhilBERTa & 0.23$^{*}$ & -- & -- \\
        \bottomrule
    \end{tabular}
    \caption{\textit{Odes} Macro-F\textsubscript{1} scores for models trained with data annotated with PC clustering. Since two loss functions were applied per embedding-encoder pair, we show only each pair's maximum score. Values with a $*$ use cross entropy loss, whereas values with a $\dagger$ use GDW-CE loss.}
    \label{tab:coordinate-results}
\end{table}

\subsection{Experimental Design}
\label{sec:experimental-design}

We divided our annotated data into three splits for training, validation, and testing. The sets contained 80\% (61,204 examples), 10\% (7,651 examples), and 10\% (7,650 examples) of the overall data, respectively. We used the validation data during training to permit early stopping, setting Macro-F\textsubscript{1} as our criterion of interest with a patience of 10. Otherwise, training would halt after 100 epochs.

We implemented our neural architecture with the PyTorch library \cite{paszkePyTorchImperativeStyle2019}. With a fixed random seed, model inputs were tokenized and truncated to the maximum sequence length of the selected Latin LM. They were grouped into batches of size 16 for all LMs save for CANINE-C and CANINE-S, as such models stressed memory resources with a maximum sequence length of 2048; in this case, we used a batch size of 8. 

When Transformers were used, we fixed their attention heads to 8, used ReLU activations, and applied PreNorm \cite{chenBestBothWorlds2018,nguyenTransformersTearsImproving2019}. We used either cross-entropy or GDW-CE to compute the loss. We optimized the neural networks with the Adam optimizer \cite{kingmaAdamMethodStochastic2015}, and gradients were clipped with an L\textsubscript{2} norm of 1 \cite{pascanuDifficultyTrainingRecurrent2013}.

\subsection{Hyperparameter Search}
\label{sec:hyperparameter-search}

To avoid falling prey to poor hyperparameter selections for each instance of our architecture, we perform a random hyperparameter search \cite{bergstraRandomSearchHyperparameter2012} of four trials for each instance. We vary the learning rate, hidden size, and number of layers in the encoder. We provide the ranges for these values with this work's repository.

Instances were constructed by fixing four modeling components: the embedding, the encoder, the dataset, and the loss function. SPhilBERTa was only employed with the PC dataset, as it was used to create the Gaussian dataset; moreover, it only used the identity-based encoder, as it creates a sequence-level embedding. Finally, the GDW-CE loss was only applied with the PC dataset. 

Once all sets of four trials were finished, we evaluated these models on the automatically-annotated test set. The best model among these four was then tested on the \textit{Odes} data. 

\begin{table}[t]
    \small
    \centering
    \begin{tabular}{lrrr}
        \toprule
        \multirow[t]{2}{5em}{\textbf{Embedding}} & \multicolumn{3}{c}{\textbf{Encoder}} \\
        & Identity & LSTM & Transformer \\
        \midrule
        Latin BERT & 0.38 & 0.38 & 0.38 \\
        LaBERTa & 0.31 & 0.31 & 0.37 \\
        PhilBERTa & 0.24 & \textbf{0.39} & \textbf{0.41} \\
        mBERT & 0.19 & 0.20 & 0.30 \\
        CANINE-C & 0.26 & 0.33 & 0.24 \\
        CANINE-S & 0.27 & 0.37 & 0.30 \\
        \bottomrule
    \end{tabular}
    \caption{\textit{Odes} Macro-F\textsubscript{1} scores for models trained with data annotated with Gaussian clustering.}
    \label{tab:gaussian-results}
\end{table}

\section{Results}
\label{sec:results}

\begin{figure*}[t]
    \centering
    \begin{tabular}{l|rr|r}
        \toprule
        \multirow[t]{2}{3em}{\textbf{Split}} & \multicolumn{2}{c}{Macro-Avg.} & Micro-Avg. \\
        & Score ($\uparrow$) & Rank ($\downarrow$) & Score ($\uparrow$) \\
        \midrule
        Horace & 0.29 & 3 & -- \\
        Pontano & 0.42 & 1 & -- \\
        Seneca & 0.12 & 4 & -- \\
        \midrule[.1em]
        Total & 0.28 & 2 & 0.22 \\
        \bottomrule
    \end{tabular}
    \hspace{1ex}
    \begin{tabular}{l|rr|r}
        \toprule
        \multirow[t]{2}{3em}{\textbf{Split}} & \multicolumn{2}{c}{Macro-Avg.} & Micro-Avg. \\
        & Score ($\uparrow$) & Rank ($\downarrow$) & Score ($\uparrow$) \\
        \midrule
        Horace & 0.21 & 4 & -- \\
        Pontano & 0.31 & 3 & -- \\
        Seneca & 0.14 & 3 & -- \\
        \midrule[.1em]
        Total & 0.22 & 4 & 0.22 \\
        \bottomrule
    \end{tabular}
    \caption{Ranks and reported Macro-F\textsubscript{1} score averages for our EvaLatin 2024 shared task submissions. The left and right tables are for the first and second submissions, respectively. Ranks range between 1 and 4, not accounting for the baseline. When a tie occurs, the best possible ranking is displayed.}
    \label{fig:submission-scores}
\end{figure*}

\begin{figure*}[t]
    \centering
    \begin{subfigure}{.45\textwidth}
    \includegraphics[width=\linewidth]{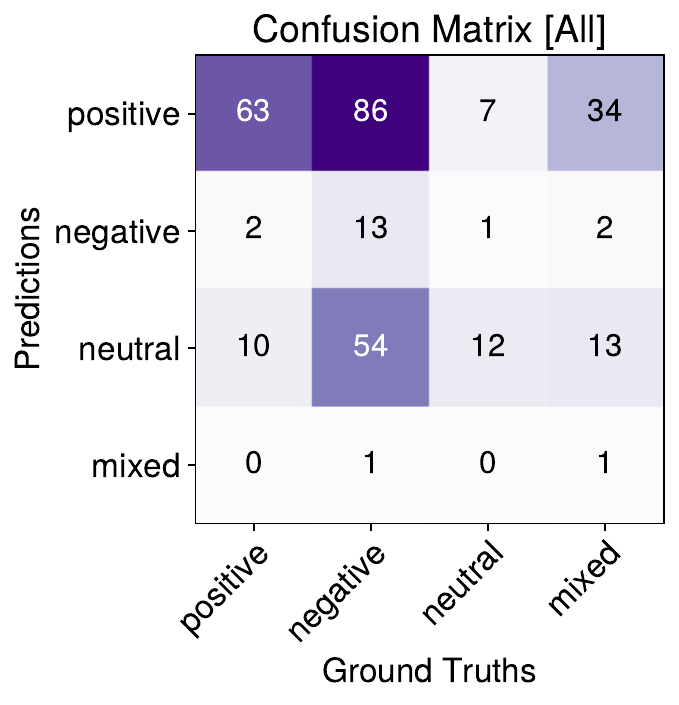}
    \end{subfigure}
    \hspace{1ex}
    \begin{subfigure}{.45\textwidth}
    \includegraphics[width=\linewidth]{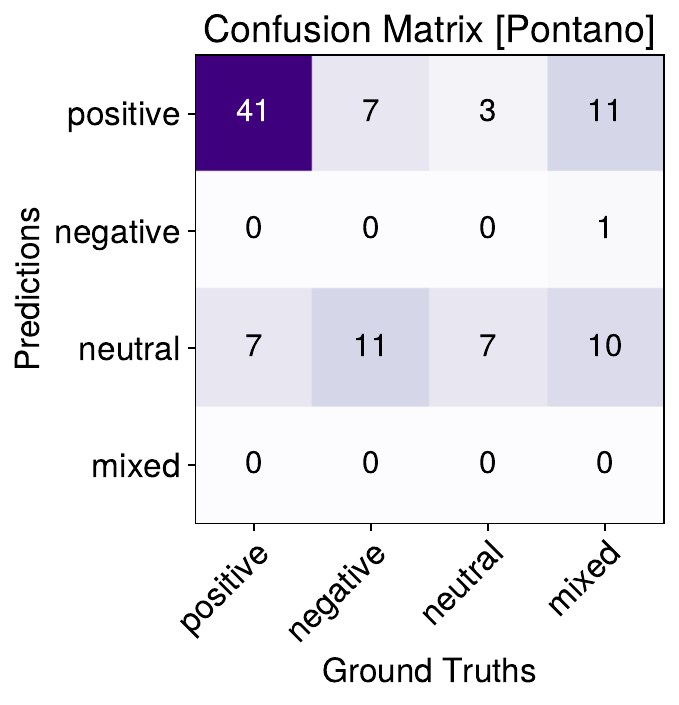}
    \end{subfigure}
    \caption{Confusion matrices for our best-performing submission. The left matrix is for the whole EvaLatin 2024 test set, whereas the right matrix is for the Pontano subset. Darker colors indicate larger values on the heatmap; text colors are shifted for readability.}
    \label{fig:best-confusion-matrices}
\end{figure*}

We present a sampling of our experimental results in \Cref{tab:coordinate-results,tab:gaussian-results}, emboldening top two results across both tables. 
According to the \textit{Odes} test set, the Gaussian dataset had a more reliable signal for sentiment. Our top two results used PhilBERTa embeddings with non-identity encoders. We submitted these models to the shared task, labeling the Transformer encoder model as our first submission and the BiLSTM encoder model as our second. 

We provide our results in the shared task in \cref{fig:submission-scores}. The first submission generally outperformed the second, only falling below the other on our worst-performing split: Seneca's \textit{Hercules Furens}. When considering other teams' submissions, our first submission achieved the best macro-averaged Macro-F\textsubscript{1} score on the Pontano split by 0.1 points, and it narrowly missed tying for the top overall score (merited by TartuNLP) by 0.01 points. Thus, although our method did not place first, it nevertheless closely rivaled the best-performing method.


One question arising from our results concerns why the Gaussian dataset broadly outperformed the PC dataset. We speculate that this relates to the distributions of the underlying data, as presented in \Cref{tab:dataset-distributions}. The PC dataset heavily favored the neutral class; whether this resembles the true distribution or not, it poorly matched the distributions of the test set. The neutral class is consistently the smallest class among the  emotionally-charged poems (Horace), lullabies (Pontano), and tragedy (Seneca) in the test set. Conversely, the Gaussian dataset has a more balanced spread of classes. Yet the lean of the Gaussian dataset's distribution into the positive class may help to explain our model's first-place performance on the Pontano subset.

To provide further evidence for this claim, we depict confusion matrices for our best-performing submission in \Cref{fig:best-confusion-matrices}.\footnote{The  matrices for our other submission are quite similar, so the trends described also apply to it.} For both the whole test set and the Pontano subset, the model primarily predicted the positive class, followed by the neutral class. In the case of the full dataset, these positive guesses add up to the largest sources of error: the model frequently mistakes negative sentences for positive ones. This effect is drastically reduced in the Pontano subset, as most of the sentences are positive. Altogether, these points further signal the meaningful influence of the Gaussian dataset's distribution on the model's performance. 

To examine this influence in more detail, we check the level of agreement between our best neural models and the original Gaussian clustering annotator. Running EvaLatin's test data through the Gaussian model, 
we use Cohen's $\kappa$ \cite{cohen1960coefficient,artstein2008inter} to measure our models' agreement beyond chance. Our top two neural models, which were trained on the Gaussian model's automatically annotated data, have $\kappa$ values of $0.32$ and $0.38$. These weak agreement scores in combination with the prior evidence seem to imply that, although the neural models roughly inherited the Gaussian annotator's classification distribution, the networks' additional learning produced distinct cues for classification labels. Such effects may be ripe material for further investigation in improving low-resource polarity detection.



\section{Conclusion}
\label{sec:conclusion}

This paper presents two methods for data augmentation in a low-resource context. Each method employs a clustering-based approach to automatically annotate Latin data for polarity detection. The best of our models, using PhilBERTa-based embeddings, a Transformer encoder, and our dataset derived from Gaussian clustering,  placed second in the task based on the macro-averaged Macro-F\textsubscript{1} score. Future work could explore the refinement of automatically-annotated data, perhaps integrating the expectation-maximization style of Gaussian training into a neural network to account for noise.

\section{Acknowledgements}

We would like to thank the EvaLatin organizers for their work in facilitating this shared task. We would also like to thank Margherita Fantoli for providing some clarifications about the Archimedes Latinus treebank \citeplanguageresource{fantoliLinguisticAnnotationNeoLatin2022}. Finally, we would like to thank Darcey Riley, Ken Sible, Aarohi Srivastava, Chihiro Taguchi, Andy Yang, and Walter Scheirer for their feedback and support.

This research was supported in part by an FRSP grant from the University of Notre Dame.




\nocite{*}
\section{Bibliographical References}
\label{sec:reference}
\vspace*{-4ex}
\bibliographystyle{lrec-coling2024-natbib}
\bibliography{evalatin}

\section{Language Resource References}
\label{lr:ref}
\vspace*{-4ex}
\bibliographystylelanguageresource{lrec-coling2024-natbib}
\bibliographylanguageresource{evalatin-langresource}

\end{document}